\newcommand{\CLASSINPUTtoptextmargin}{1.91cm}
\newcommand{\CLASSINPUTbottomtextmargin}{2.5cm}
\documentclass[10pt, conference]{IEEEtran}
\usepackage{amsmath, amssymb, amsfonts, amssymb, amsthm}
\usepackage{bbm}
\usepackage{graphicx} \graphicspath{ {figure/} }
\usepackage{epstopdf} 
\usepackage[noabbrev]{cleveref}
\usepackage{scalefnt}
\usepackage{indentfirst}
\usepackage{cite}
\usepackage{bm}
\usepackage{enumerate}
\usepackage{enumitem}
\usepackage[caption=false,subrefformat=parens,labelformat=parens]{subfig}
\usepackage{scalerel}
\usepackage{scalerel}
\usepackage{algpseudocode}
\usepackage{etoolbox}
\usepackage{multirow}
\usepackage[table,xcdraw]{xcolor}

\usepackage{algorithm}
\usepackage{algcompatible}
\renewcommand{\algorithmiccomment}[1]{\hfill ~#1}
\algnewcommand\INPUT{\item[\textbf{Input:}]}
\algnewcommand\INITIAL{\item[\textbf{Initialization:}]}
\algnewcommand\OUTPUT{\item[\textbf{Output:}]}
\algnewcommand\RETURN{\item[\textbf{Return:}]}
\algnewcommand\ITER{\item[\textbf{Iteration:}]}
\algrenewcommand\algorithmiccomment[2][\small]{{#1\hfill\ #2}}

\theoremstyle{plain}

\makeatletter
\newcommand*{\algrule}[1][\algorithmicindent]{%
  \makebox[#1][l]{%
    \hspace*{.2em}% <------------- This is where the rule starts from
    \vrule height .75\baselineskip depth .25\baselineskip
  }
}

\newcount\ALG@printindent@tempcnta
\def\ALG@printindent{%
    \ifnum \theALG@nested>0% is there anything to print
    \ifx\ALG@text\ALG@x@notext% is this an end group without any text?
    \else
    \unskip
    \ALG@printindent@tempcnta=1
    \loop
    \algrule[\csname ALG@ind@\the\ALG@printindent@tempcnta\endcsname]%
    \advance \ALG@printindent@tempcnta 1
    \ifnum \ALG@printindent@tempcnta<\numexpr\theALG@nested+1\relax
    \repeat
    \fi
    \fi
}

\patchcmd{\ALG@doentity}{\noindent\hskip\ALG@tlm}{\ALG@printindent}{}{\errmessage{failed to patch}}
\patchcmd{\ALG@doentity}{\item[]\nointerlineskip}{}{}{} % no spurious vertical space
\makeatother

\renewcommand{\algorithmiccomment}[2][.5\linewidth]{\leavevmode\hfill\makebox[#1][l]{//~#2}}

\theoremstyle{definition}

\begin{document}
\title{Deep Learning and Matrix Completion-aided IoT Network Localization in the Outlier Scenarios}

\author{
    \IEEEauthorblockN{Sunwoo Kim}
    \IEEEauthorblockA{Department of Electrical and Computer Engineering, Seoul National University, Korea
    \\swkim@islab.snu.ac.kr}
    
}

\maketitle

\begin{abstract}
In this paper, we propose a deep learning and matrix completion aided approach for recovering  an outlier contaminated Euclidean distance matrix D in IoT network localization.
Unlike conventional localization techniques that search the solution over a whole set of matrices, the proposed technique restricts the search to the set of Euclidean distance matrices. Specifically, we express D as a function of the sensor coordinate matrix X that inherently satisfies the unique properties of D, and then jointly recover D and X using a deep neural network.
To handle outliers effectively, we model them as a sparse matrix L and add a regularization term of L into the optimization problem. We then solve the problem by alternately updating X, D, and L. Numerical experiments demonstrate that the proposed technique can recover the location information of sensors accurately even in the presence of outliers.
\end{abstract}

% \begin{IEEEkeywords}
% mobility management, handover, terahertz communications, beamforming, ultra-dense network, sensing, computer vision, object detection, deep learning
% \end{IEEEkeywords}

% \newpage
\vspace{-0.1cm}
\section{Introduction}
With the advent of the Internet of Things (IoT) era, massive machine-type communication (mMTC) has gained significant attention as one of the key use cases for B5G and 6G~\cite{mmtc}. The primary goal of mMTC is to support massive connectivity for a large number of sensors to the data center (e.g.,  eNB/gNB~\cite{globecom} or base station~\cite{basestation}). To achieve this goal, the data center should determine the location of all sensors~\cite{localization:rawat2014}. This process, known as \textit{network localization}, involves constructing a location map of the sensors using the pairwise distance information between each sensor pair~\cite{localization:pal2010}. For this purpose, each sensor needs to report the distances to its adjacent sensors~\cite{localization:aspnes2006}.

In practice, however, only a limited portion of the distance information is available at the data center, making network localization a challenging problem.
This scarcity of distance information is mainly caused by the limited radio communication range~\cite{localization:dong2016}, which results in many unknown entries in the observed Euclidean distance matrix $\mathbf{D_o}$.

One popular strategy to recover the original distance matrix $\mathbf{D}$ from $\mathbf{D_{o}}$ is a low-rank matrix completion (LRMC) \cite{skim_localization2}. 
Considering that the rank of the Euclidean distance matrix $\mathbf{D}$ in the $k$-dimensional Euclidean space is at most $k+2$ (e.g., $\text{rank}$ is $5$ for $3$-dimensional location map) \cite{skim_localization2}, it can be modeled as a low-rank matrix. The problem to recover a low-rank matrix $\mathbf{D}$ from $\mathbf{D_{o}}$ can be expressed as
\begin{equation} \label{eq:original problem}
    \begin{split}
        &\min_{\widehat{\mathbf{D}} \in \mathbb{R}^{n\times n}} ~~~~ \|\mathcal{P}_{E}(\widehat{\mathbf{D}})-\mathcal{P}_{E}(\mathbf{D_{o}})\|_{F}^{2} \\
        &~~~\hspace{.5mm}\text{s.t.}~~~~~~~\text{rank}(\widehat{\mathbf{D}}) \leq k+2,
    \end{split}
\end{equation}
where $\mathcal{P}_E$ is the sampling operator given by 
\begin{equation}
[\mathcal{P}_E(\mathbf{A})]_{ij} = \left\lbrace\begin{matrix}
A_{ij} & \text{if } (i,j)\in E\\
0 & \text{otherwise}.
\end{matrix} \right. 
\label{eq:eq210}
\end{equation}  
Note here that $E$ is the set of observed indices for a given radio communication range $r$. Over the years, various LRMC techniques~\cite{localization:wen2012, localization:hu2013, localization:vandereycken2013} have been employed to solve \eqref{eq:original problem}.  %In 

To solve~\eqref{eq:original problem} effectively, a deep learning (DL)-based LRMC technique has been proposed recently~\cite{skim_localization}.
The core idea
 of this technique is to estimate $\mathbf{D}$ over the set of Euclidean distance matrices.
 To this end, DL-based LRMC expresses $\mathbf{D}$ as a function $g(\mathbf{X})$ of the sensor coordinate matrix $\mathbf{X}$ that inherently ensures the symmetry and zero diagonal and positive non-diagonal entries of $\mathbf{D}$, and then jointly recovers $\mathbf{D}$ and $\mathbf{X}$ satisfying $\mathbf{D} = g(\mathbf{X})$ and $\mathcal{P}_{E}(\mathbf{D}) = \mathcal{P}_{E}(\mathbf{D_{o}})$ as
\begin{equation} \label{eq:main problem_intro}
    \begin{split}
        &\min_{\widehat{\mathbf{X}} \:\in\:\mathbb{R}^{n \times k}, \widehat{\mathbf{D}} \:\in\:\mathbb{R}^{n\times n}} ~~~~ \|g(\widehat{\mathbf{X}})-\widehat{\mathbf{D}}\|_F^2, \\
        &~~~~~~~~~\hspace{.5mm}\text{s.t.}~~~~~~~~~~~~~\mathcal{P}_{E}(\widehat{\mathbf{D}}) = \mathcal{P}_{E}(\mathbf{D_{o}}).
    \end{split}
\end{equation}
 Efficacy of this technique depends heavily on the
 quality of observed pairwise distances but conventional DL-based LRMC technique cannot properly handle unwanted outliers, which contaminate observed distances and occur in many practical scenarios due to various reasons including the hardware (Tx/Rx) malfunction and adversary attacks~\cite{outlier}.
 Since such outliers might degrade the localization performance severely, it must be controlled properly in the recovery process.

An aim of this paper is to propose a novel IoT network localization
technique for the outlier scenarios using DL and LRMC. 
In the proposed technique, referred to as extended multiple deep neural networks for localization (E-MDNL), we model 
the observed distance as $d^{o}_{ij} = d_{ij} + l_{ij}$
($l_{ij}$ is the outlier) and
thus $\mathcal{P}_{E}(\mathbf{D_{o}}) = \mathcal{P}_{E}(\widehat{\mathbf{D}} + \mathbf{L})$, where $\mathbf{L}$ is the outlier matrix.
Since the outlier occurs rarely, we model $\mathbf{L}$ as a sparse
matrix. 
We then add a regularization term of $\mathbf{L}$ into the problem in \eqref{eq:main problem_intro} as
\begin{align} \label{eq:main problem_introw}
&\min_{\widehat{\mathbf{D}} ,\widehat{\mathbf{X}} , \mathbf{L}\:\in\:\mathbb{R}^{n\times n}} ~~~~ \|g(\widehat{\mathbf{X}})-(\widehat{\mathbf{D}} + \mathbf{L}) \|_F^2 + \tau \| \mathbf{L} \|_{1}, \nonumber \\ &~~~~~~~~~\hspace{.5mm}\text{s.t.}~~~~~~~~~~~~~\mathcal{P}_{E}(\widehat{\mathbf{D}} +\mathbf{L}) = \mathcal{P}_{E}(\mathbf{D_{o}}),
\end{align} 
where $\tau$ is the regularization factor that controls the trade-off between the sparsity of $\mathbf{L}$ and the consistency of the observed pairwise distances.

In solving \eqref{eq:main problem_introw}, E-MDNL updates $\widehat{\mathbf{D}}$, $\widehat{\mathbf{X}}$, and $\mathbf{L}$ alternately while fixing the other estimates. For given $\widehat{\mathbf{X}}$ and $\mathbf{L}$ (or $\widehat{\mathbf{X}}$ and $\mathbf{D}$), we analytically derive a closed-form solution of \eqref{eq:main problem_introw}, using which E-MDNL updates $\widehat{\mathbf{D}}$ (or $\mathbf{L}$) efficiently. 
When $\widehat{\mathbf{D}}$ and $\mathbf{L}$ are fixed, it is not easy to find out a closed-form solution to \eqref{eq:main problem_introw}. 
To identify a nonlinear and complicated mapping from $\widehat{\mathbf{D}}$ to $\widehat{\mathbf{X}}$, E-MDNL uses a deep neural network (DNN), a powerful tool of DL that can well approximate any given function~\cite{skim_localization2}. 

From the simulation results, we
show that E-MDNL can accurately recover the outlier-contaminated Euclidean distance matrix via DNN and LRMC. 

\section{Proposed E-MDNL Technique}\label{sec:2}

In this work, we consider the scenario where $n$ sensor nodes are distributed in the $k$-dimensional Euclidean space. Let $\mathbf{x}_{i} \in \mathbb{R}^{k}$ be the $i$-th sensor's  coordinate vector and $d_{ij}$ be the distance between the $i$-th and $j$-th sensors. Then, $d_{ij}^{2} = \| \mathbf{x}_{i} - \mathbf{x}_{j} \|_{2}^{2} = \mathbf{x}_i^T\mathbf{x}_i + \mathbf{x}_j^T\mathbf{x}_j - 2\mathbf{x}_i^T\mathbf{x}_j$, and thus we have
 \begin{align} \label{eq:relationship between D and X}
     \mathbf{D}
     &= \mathbf{1}\text{diag}(\mathbf{XX}^T)^T + \text{diag}(\mathbf{XX}^T)\mathbf{1}^T - 2\mathbf{XX}^T,
 \end{align}
 where $\mathbf{X} = [ \mathbf{x}_{1} \ \mathbf{x}_{2} \ \cdots \ \mathbf{x}_{n} ]^{T} \in \mathbb{R}^{n \times k}$ is the sensor coordinate matrix. 

\begin{figure*}[t]
\centering
{\includegraphics[width=2.0\columnwidth, height=9.0cm]{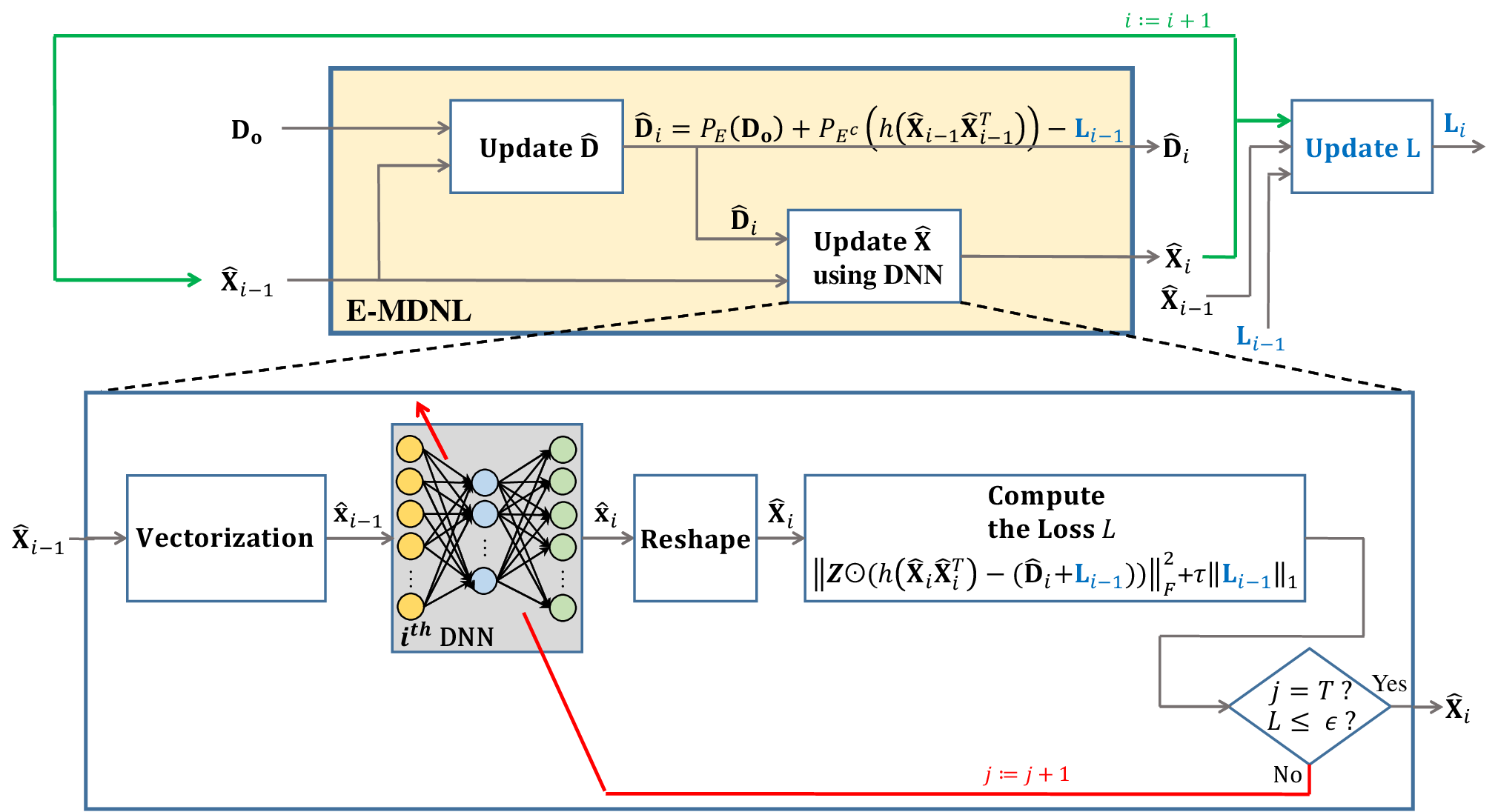}}
	\caption{Overall process of the proposed E-MDNL technique. }
	\label{fig:datasetselectioncode}
 \vspace{-0.5cm}
\end{figure*}
Overall process of the proposed E-MDNL technique is as follows. To find out $\mathbf{X}$ and $\mathbf{D}$ satisfying~\eqref{eq:relationship between D and X} and $\mathcal{P}_{E}(\mathbf{D} + \mathbf{L}) = \mathcal{P}_{E}(\mathbf{D_{o}})$, E-MDNL solves the minimization problem
  \begin{align}
       &\min_{\widehat{\mathbf{D}} , \widehat{\mathbf{X}}, \mathbf{L}\:\in\:\mathbb{R}^{n\times n}} ~~~~ \|h(\widehat{\mathbf{X}}\widehat{\mathbf{X}}^T)-(\widehat{\mathbf{D}} + \mathbf{L}) \|_F^2 + \tau \| \mathbf{L} \|_{1}, \nonumber \\ &~~~~~~~~~\hspace{.5mm}\text{s.t.}~~~~~~~~~~~~~\mathcal{P}_{E}(\widehat{\mathbf{D}} +\mathbf{L}) = \mathcal{P}_{E}(\mathbf{D_{o}}).
\end{align} 
 by updating $\widehat{\mathbf{X}}$, $\widehat{\mathbf{D}}$, and $\mathbf{L}$ alternately, where $h(\widehat{\mathbf{X}} \widehat{\mathbf{X}}^{T}) = \mathbf{1} \text{diag}(\widehat{\mathbf{X}} \widehat{\mathbf{X}}^{T})^{T} + \text{diag}(\widehat{\mathbf{X}} \widehat{\mathbf{X}}^{T}) \mathbf{1}^{T} - 2 \widehat{\mathbf{X}} \widehat{\mathbf{X}}^{T}$.
 Starting from a randomly initialized matrix $\widehat{\mathbf{X}}_{0}$ and a matrix full of zeros $\mathbf{L}_{0}$, E-MDNL updates the $i$-th estimates $\widehat{\mathbf{D}}_{i}$, $\widehat{\mathbf{X}}_{i}$, and $\mathbf{L}_{i}$ by alternately solving the following three subproblems 
 \begin{equation} \label{eq:update of D}
 \begin{split}
 &\min_{\widehat{\mathbf{D}} \:\in\:\mathbb{R}^{n\times n}} ~~~~ \|h(\widehat{\mathbf{X}}_{i-1} \widehat{\mathbf{X}}_{i-1}^{T})-(\widehat{\mathbf{D}} + \mathbf{L}_{i-1}) \|_{F}^{2} + \tau \| \mathbf{L}_{i-1} \|_{1}, \\
 &~~~~\hspace{.5mm}\text{s.t.}~~~~~~~~\mathcal{P}_{E}(\widehat{\mathbf{D}} + \mathbf{L}_{i-1}) = \mathcal{P}_{E}(\mathbf{D_{o}}),
 \end{split}
 \end{equation}
 and
 \begin{align} \label{eq:update of X}
     \min_{\widehat{\mathbf{X}} \in \mathbb{R}^{n \times k}} ~~~~ \| h(\widehat{\mathbf{X}} \widehat{\mathbf{X}}^{T})-(\widehat{\mathbf{D}}_{i} + \mathbf{L}_{i-1}) \|_{F}^{2} + \tau \| \mathbf{L}_{i-1} \|_{1},
 \end{align}
 and
 \begin{equation} \label{eq:update of L}
 \begin{split}
 &\min_{\mathbf{L} \:\in\:\mathbb{R}^{n\times n}} ~~~~ \|h(\widehat{\mathbf{X}}_{i} \widehat{\mathbf{X}}_{i}^{T})-(\widehat{\mathbf{D}}_{i} + \mathbf{L})\|_{F}^{2} +  \tau \| \mathbf{L} \|_{1}, \\
 &~~~~\hspace{.5mm}\text{s.t.}~~~~~~~~\mathcal{P}_{E}(\widehat{\mathbf{D}}_{i} + \mathbf{L}) = \mathcal{P}_{E}(\mathbf{D_{o}}),
 \end{split}
 \end{equation}

 \subsection{Euclidean Distance Matrix Estimate $\widehat{\mathbf{D}}$ Update} \label{sec:update of D}
 
 In order to update the estimate $\widehat{\mathbf{D}}$ of the Euclidean distance matrix when the sensor coordinate matrix estimate $\widehat{\mathbf{X}}_{i-1}$ and  $\mathbf{L}_{i-1}$ are given, E-MDNL finds out the solution to the subproblem in~\eqref{eq:update of D}. For any given matrix $\mathbf{M} \in \mathbb{R}^{n \times n}$, we have
 \begin{align}
     \| \mathbf{M} \|_{F}^{2}
     \hspace{-.1cm} &= \hspace{-.05cm} \| \mathcal{P}_{E}(\mathbf{M}) \hspace{-.05cm} + \hspace{-.05cm} \mathcal{P}_{E^{c}}(\mathbf{M}) \|_{F}^{2} \nonumber \\
     \hspace{-.1cm} &= \hspace{-.05cm} \|  \mathcal{P}_{E}(\mathbf{M}) \|_{F}^{2} \hspace{-.05cm} + \hspace{-.05cm} \|  \mathcal{P}_{E^{c}} (\mathbf{M}) \|_{F}^{2}  \nonumber \\
     \hspace{-.1cm} &+ 2 \text{tr}(\mathcal{P}_{E}(\mathbf{M})^{T} \mathcal{P}_{E^{c}} (\mathbf{M}))  \nonumber \\
     & \hspace{-.05cm} \overset{(a)}{=} \hspace{-.05cm} \|  \mathcal{P}_{E}(\mathbf{M}) \|_{F}^{2} \hspace{-.05cm} + \hspace{-.05cm} \|  \mathcal{P}_{E^{c}} (\mathbf{M}) \|_{F}^{2}, \label{eq:separation}
 \end{align}
where (a) follows from $\text{tr}(\mathcal{P}_{E}(\mathbf{M})^{T} \mathcal{P}_{E^{c}}(\mathbf{M})) = 0$.\footnote{Note that $\text{tr}(\mathcal{P}_{E}(\mathbf{M})^{T} \mathcal{P}_{E^{c}} (\mathbf{M}))  = \sum_{k=1}^{n} \mathcal{P}_{E}(\mathbf{M})_{ki} \mathcal{P}_{E^{c}} (\mathbf{M})_{ki} = 0$ since either $\mathcal{P}_{E}(\mathbf{M})_{ki}$ or $\mathcal{P}_{E^{c}} (\mathbf{M})_{ki}$ would be zero.} Using this, the objective function in~\eqref{eq:update of D} can be expressed as
\begin{align}
     \lefteqn{ \|h(\widehat{\mathbf{X}}_{i-1} \widehat{\mathbf{X}}_{i-1}^{T})-(\widehat{\mathbf{D}} + \mathbf{L}_{i-1})\|_{F}^{2} + \tau \| \mathbf{L}_{i-1} \|_{1} } \nonumber \\
    &\hspace{-.1cm} \overset{(b)}{=} \hspace{-.1cm}  \|\mathcal{P}_{E}
    (h(\widehat{\mathbf{X}}_{i-1} \widehat{\mathbf{X}}_{i-1}^{T})
    \hspace{-.1cm} - \hspace{-.1cm} (\widehat{\mathbf{D}} + \mathbf{L}_{i-1}))\|_F^2  \\
    &+ \| \mathcal{P}_{E^c}(h(\widehat{\mathbf{X}}_{i - 1}\widehat{\mathbf{X}}_{i - 1}^T)\hspace{-.1cm} - \hspace{-.1cm} (\widehat{\mathbf{D}} + \mathbf{L}_{i-1}))\|_F^2 + \tau \| \mathbf{L}_{i-1} \|_{1},\nonumber \nonumber\\
    &\hspace{-.1cm}\overset{(c)}{=} \hspace{-.1cm}  \| \mathcal{P}_{E}(h(\widehat{\mathbf{X}}_{i - 1}\widehat{\mathbf{X}}_{i - 1}^T) \hspace{-.1cm} - \hspace{-.1cm} \mathbf{D_o})\|_F^2 \hspace{-.1cm} \nonumber \\
    &+ \| \mathcal{P}_{E^c}(h(\widehat{\mathbf{X}}_{i - 1}\widehat{\mathbf{X}}_{i - 1}^T) \hspace{-.1cm} - \hspace{-.1cm} (\widehat{\mathbf{D}} + \mathbf{L}_{i-1}))\|_F^2 + \tau \| \mathbf{L}_{i-1} \|_{1}, \nonumber
\end{align}
where (b) is from~\eqref{eq:separation} and (c) is because $\mathcal{P}_{E}(\widehat{\mathbf{D}} + \mathbf{L}_{i-1}) = \mathcal{P}_{E}(\mathbf{D_{o}})$. Since $\|\mathcal{P}_{E}(h(\widehat{\mathbf{X}}_{i - 1}\widehat{\mathbf{X}}_{i - 1}^T) - \mathbf{D_o})\|_F^2$ and $\tau \|\mathbf{L}_{i-1} \|_{1}$ are independent of $\widehat{\mathbf{D}}$, the minimization problem in~\eqref{eq:update of D} can be reformulated as
 \begin{equation} \label{eq:modified update of D}
 \begin{split}
 &\min_{\widehat{\mathbf{D}} \:\in\:\mathbb{R}^{n\times n}} ~~~~ \|\mathcal{P}_{E^{c}}(h(\widehat{\mathbf{X}}_{i - 1}\widehat{\mathbf{X}}_{i - 1}^T) \hspace{-.1cm} - \hspace{-.1cm} (\widehat{\mathbf{D}} + \mathbf{L}_{i-1}))\|_F^2, \\
 &~~~~\hspace{.5mm}\text{s.t.}~~~~~~~~\mathcal{P}_{E}(\widehat{\mathbf{D}} + \mathbf{L}_{i-1})) = \mathcal{P}_{E}(\mathbf{D_{o}}).
 \end{split}
 \end{equation}
 It is now clear that the objective function in~\eqref{eq:modified update of D} is minimized when
\begin{align}
    & \mathcal{P}_{E^c}(\widehat{\mathbf{D}} + \mathbf{L}_{i-1}) = \mathcal{P}_{E^c}(h(\widehat{\mathbf{X}}_{i - 1}\widehat{\mathbf{X}}_{i - 1}^T)), \\
    & \mathcal{P}_E(\widehat{\mathbf{D}} + \mathbf{L}_{i-1})  = \mathcal{P}_E(\mathbf{D_{o}})
 \end{align}
 Consequently, the proposed E-MDNL technique updates the Euclidean distance matrix estimate $\widehat{\mathbf{D}}_{i}$ as
 \begin{align}
    \widehat{\mathbf{D}}_{i} +  \mathbf{L}_{i-1} 
    &= \mathcal{P}_{E}(\widehat{\mathbf{D}}_{i} + \mathbf{L}_{i-1}) + \mathcal{P}_{E^c}(\widehat{\mathbf{D}}_{i} + \mathbf{L}_{i-1}) \nonumber\\
    &= \mathcal{P}_E(\mathbf{D_{o}}) + \mathcal{P}_{E^c}(h(\widehat{\mathbf{X}}_{i - 1}\widehat{\mathbf{X}}_{i - 1}^T))\nonumber\\
    \widehat{\mathbf{D}}_{i} &= 
    \mathcal{P}_E(\mathbf{D_{o}}) + \mathcal{P}_{E^c}(h(\widehat{\mathbf{X}}_{i - 1}\widehat{\mathbf{X}}_{i - 1}^T)) - \mathbf{L}_{i-1} 
    \label{eq:update formula of D}
 \end{align}
 
\subsection{Sensor Coordinate Matrix Estimate $\widehat{\mathbf{X}}$ Update}
 
\label{sec:update of X}
Once $\widehat{\mathbf{D}}_{i}$ is obtained, E-MDNL updates the sensor coordinate matrix estimate $\widehat{\mathbf{X}}$ by solving~\eqref{eq:update of X}. Since it is difficult to obtain the closed-form solution of~\eqref{eq:update of X}, we use the DNN, an efficient tool to approximate a nonlinear, complicated function~\cite{skim_localization2}.

The proposed neural network of E-MDNL consists of $m$ fully-connected (FC) layers, $m-1$ hidden layers, and one output layer. Let $(\mathbf{W}_{il}, \mathbf{b}_{il})$ and $(\mathbf{W}_{iz}, \mathbf{b}_{iz})$ be the weight and bias parameters of the $\ell$-th hidden layer and the output layer, respectively, then the output $\widehat{\mathbf{x}}_{i}$ of the last layer is defined as
\begin{equation}
    \widehat{\mathbf{x}}_{i} = \mathbf{W}_{iz} \mathbf{h}_{i,m-1} + \mathbf{b}_{iz},
\end{equation}
where $\mathbf{h}_{i, \ell}=\sigma(\mathbf{W}_{i\ell} \mathbf{h}_{i, \ell-1} + \mathbf{b}_{i\ell})$ is the output of the $\ell$-th hidden layer $(1 \le \ell \le m-1)$, $\mathbf{h}_{i, 0} = \text{vec}(\widehat{\mathbf{X}}_{i-1})$, and $\sigma$ is a pre-defined nonlinear activation function. Basically, the output $\widehat{\mathbf{x}}_{i}$ of the DNN can be expressed as
\begin{align}
    \widehat{\mathbf{x}}_{i}
    &= f(\widehat{\mathbf{X}}_{i-1}; \Theta),
\end{align}
where $\Theta = \{ \mathbf{W}_{i\ell}, \mathbf{W}_{iz}, \mathbf{b}_{i \ell}, \mathbf{b}_{iz} : 1 \le \ell \le m-1 \}$ is the set of network parameters of the DNN.

Since the main goal of the DNN in E-MDNL is to find out the solution of~\eqref{eq:update of X}, we iteratively update the network parameters (weight and bias) in the direction of minimizing the loss function 
\begin{align*}
       L = \|(h(\widehat{\mathbf{X}} \widehat{\mathbf{X}}^{T})-(\widehat{\mathbf{D}}_{i} + \mathbf{L}_{i-1})) \|_{F}^{2} + \tau \| \mathbf{L}_{i-1} \|_{1},
    \label{eq:loss function}
\end{align*}
where $\widehat{\mathbf{X}}_{i}$ is the matrix form of the output $\widehat{\mathbf{x}}_{i} = f(\widehat{\mathbf{X}}_{i-1}; \Theta)$.
As a tool to update the network parameters, one can use the stochastic gradient descent (SGD) method, an algorithm to find network parameters minimizing the loss function. 
This update process is repeated until the iteration number $j$ reaches the pre-defined maximum value $T$.
%}

 \subsection{Outlier Matrix  $\mathbf{L}$ Update} \label{sec:update of L}
Finally, after obtaining $\widehat{\mathbf{D}}_{i}$ and $\widehat{\mathbf{X}}_{i}$, E-MDNL updates the outlier matrix $\mathbf{L}$ by solving~\eqref{eq:update of L}.
The objective function in the third subproblem can be expressed as
    \begin{align}
    \lefteqn{ \| (h(\widehat{\mathbf{X}}_{i} \widehat{\mathbf{X}}_{i}^{T})-(\widehat{\mathbf{D}}_{i} + \mathbf{L}))\|_{F}^{2} +  \tau \| \mathbf{L} \|_{1} } \nonumber \\
    &\hspace{-.1cm}=\hspace{-.1cm} \| \mathcal{P}_{E}
    (h(\widehat{\mathbf{X}}_{i} \widehat{\mathbf{X}}_{i}^{T})
    \hspace{-.1cm} - \hspace{-.1cm} (\widehat{\mathbf{D}}_{i} + \mathbf{L}))\|_{F}^2 \\
    &+ \| \mathcal{P}_{E^c}(h(\widehat{\mathbf{X}}_{i}\widehat{\mathbf{X}}_{i}^T)\hspace{-.1cm} - \hspace{-.1cm} (\widehat{\mathbf{D}}_{i} + \mathbf{L}))\|_F^2\hspace{-.1cm}+\hspace{-.1cm}\tau \| \mathbf{L} \|_{1}\nonumber\\
    &\hspace{-.1cm}\overset{(b)}{=} \hspace{-.1cm}  \| \mathcal{P}_{E}(h(\widehat{\mathbf{X}}_{i}\widehat{\mathbf{X}}_{i}^T) \hspace{-.1cm} - \hspace{-.1cm} \mathbf{D_o})\|_F^2 \nonumber \\
    &+ \| \mathcal{P}_{E^c}(h(\widehat{\mathbf{X}}_{i}\widehat{\mathbf{X}}_{i}^T) \hspace{-.1cm} - \hspace{-.1cm} (\widehat{\mathbf{D}}_{i} + \mathbf{L}))\|_F^2 + \tau \| \mathbf{L} \|_{1}, \nonumber
    \end{align}
    where (b) is because $\mathcal{P}_{E}(\widehat{\mathbf{D}}_{i} + \mathbf{L}) = \mathcal{P}_{E}(\mathbf{D_{o}})$.
 Since $\|\mathcal{P}_{E}(h(\widehat{\mathbf{X}}_{i}\widehat{\mathbf{X}}_{i}^T) - \mathbf{D_o})\|_F^2$ is independent of $\mathbf{L}$, the third subproblem can be reformulated as
 \begin{equation*} \label{eq:modified update of L}
 \begin{split}
&\min_{\mathbf{L}\:\in\:\mathbb{R}^{n\times n}} ~~~~ \| \mathcal{P}_{E^c}(h(\widehat{\mathbf{X}}_{i }\widehat{\mathbf{X}}_{i}^T) \hspace{-.1cm} - \hspace{-.1cm} (\widehat{\mathbf{D}}_{i} + \mathbf{L}))\|_F^2 + \tau \| \mathbf{L} \|_{1}, \\
&~~~~\hspace{.5mm}\text{s.t.}~~~~~~~~\mathcal{P}_{E}(\widehat{\mathbf{D}}_{i} + \mathbf{L}) = \mathcal{P}_{E}(\mathbf{D_{o}}).
 \end{split}
 \end{equation*}
Since $\mathcal{P}_{E^c}(\widehat{\mathbf{D}}_{i} + \mathbf{L}) = \widehat{\mathbf{D}}_{i} + \mathbf{L} -\mathcal{P}_{E}(\widehat{\mathbf{D}}_{i} + \mathbf{L})$ and $\widehat{\mathbf{D}}_{i} = 
    \mathcal{P}_E(\mathbf{D_{o}}) + \mathcal{P}_{E^c}(h(\widehat{\mathbf{X}}_{i - 1}\widehat{\mathbf{X}}_{i - 1}^T)) - \mathbf{L}_{i-1}$, the objective function can be reexpressed as 
\begin{equation*}
  \| (\mathcal{P}_{E^c}(h(\widehat{\mathbf{X}}_{i }\widehat{\mathbf{X}}_{i}^T) - h(\widehat{\mathbf{X}}_{i -1}\widehat{\mathbf{X}}_{i - 1}^T)) + \mathbf{L}_{i-1} - \mathbf{L})\|_F^2 + \tau \| \mathbf{L} \|_{1}
  \end{equation*}
Thus, the problem of updating $ \mathbf{L}_{i}$ can be formulated as
 \begin{equation*} 
 \begin{split}
 \mathbf{L}_{i} &= \arg\min_{\mathbf{L}}~~ \| (\mathcal{P}_{E^c}(h(\widehat{\mathbf{X}}_{i }\widehat{\mathbf{X}}_{i}^T) - h(\widehat{\mathbf{X}}_{i -1}\widehat{\mathbf{X}}_{i - 1}^T)) \label{eq:objectfunctionL}\\
 &+ \mathbf{L}_{i-1} - \mathbf{L})\|_F^2
 + \tau \| \mathbf{L} \|_{1} \\
 &= \arg\min_{\mathbf{L}}~~ 
 \sum_{(j,k)} (c_{jk} + l_{i- 1, jk} - l_{jk})^{2} + \tau |l_{jk}|,
 \end{split}
 \end{equation*}
where $l_{i-1, jk}$, $l_{jk}$, and $c_{jk}$ are the entries of  $\mathbf{L}_{i-1}$, $\mathbf{L}$, and $\mathcal{P}_{E^c}(h(\widehat{\mathbf{X}}_{i }\widehat{\mathbf{X}}_{i}^T) - h(\widehat{\mathbf{X}}_{i -1}\widehat{\mathbf{X}}_{i - 1}^T)) $, respectively. 

To solve the problem, we use the soft-thresholding operator, which truncates the magnitude of the entries of a matrix gradually~\cite{softhresholding}. 
Using the soft-thresholding operator, the solution of (9) is
given by
 \begin{align*}
    l^{*}_{jk}
    &= \arg\min_{l_{jk}}~~ 
   (c_{jk} + l_{i- 1, jk} - l_{jk})^{2} + \tau |l_{jk}| \nonumber\\
    &=   
        \begin{cases}
        c_{jk} + l_{i- 1, jk} - \tau & \text{if}\ c_{jk} + l_{i- 1, jk} > \tau,\\
          c_{jk} + l_{i- 1, jk} + \tau & \text{if}\ c_{jk} + l_{i- 1, jk} < -\tau,\\
        0 & \text{otherwise}.\notag
            \end{cases}           
    \nonumber\\
 \end{align*}

\section{Simulations and Discussions}\label{sec:3}
\subsection{Simulation Setup}
In this section, we evaluate the performance of the proposed E-MDNL technique and compare it with the conventional MDNL technique~\cite{skim_localization}. As performance metric, we use the mean square localization error (MSLE), which is defined as
\begin{eqnarray}
MSLE &=& \frac{1}{\Gamma} \sum_{\text{All unknown nodes i}} \| \hat{\mathbf{x}}_{i} - \mathbf{x}_{i} \|_{2},\nonumber\\
\end{eqnarray}
where $\Gamma$ is the total number of unknown nodes.
$\mathbf{x}_{i}$ and $\hat{\mathbf{x}}_{i}$ are the original and reconstructed locations of
the sensor $i$, respectively.

In our simulation, we generate 200 sensors whose positions are randomly distributed in cubic space ($k = 3$), following the uniform distribution on an interval of [0,1] for 3D Euclidean space. From these sensors, a $200 \times 3$ sensor coordinate matrix 
$\mathbf{X}$ is generated and then mapped to the Euclidean distance matrix $\mathbf{D} = h(\mathbf{X}\mathbf{X}^T)$. 
We also consider the received signal strength (RSS)-based measurement model.
In this model, an entry $d_{ij}$ of $\mathbf{D_{o}}$ is known only if it is smaller than the radio communication range (i.e., $d_{ij}\leq r$)~\cite{skim_localization}. 
To model the outlier scenario, 
we randomly choose a set of the observed distances and replace this set by a set of random
numbers. 
Here it is worth mentioning that the magnitude of outliers is comparable to the distance level.
 
 In E-MDNL, we set the number of iterations $T$ to 5000 and the number of FC layers $m$ to 5. In addition, we choose ReLU function an activation function in hidden layers. 
 To update the network parameters of the DNN in E-MDNL, we employ the Adagrad optimizer, a robust gradient-based optimization tool \cite{skim_localization}. Note that we perform 1,000 independent trials and compute the average value for each point of the techniques. 

\subsection{Simulation Results}
\begin{figure}[t]
	\centering
    \includegraphics[width=1.0\columnwidth, height=6.2cm]{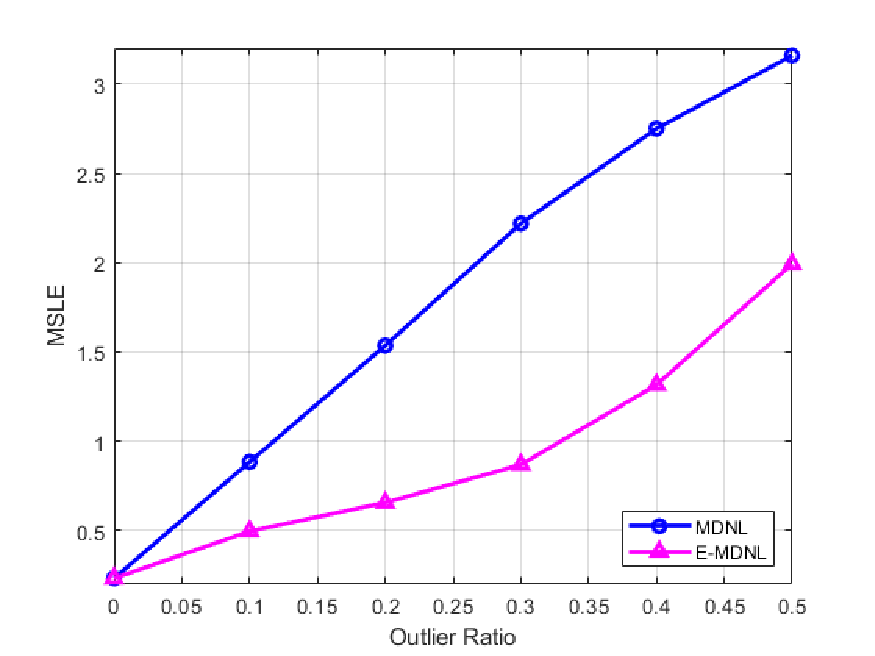}
       \vspace{-0.7cm}
	\caption{The MSLE performance of E-MDNL in the presence of outliers.}
\label{fig:result}
  \vspace{-0.4cm}
\end{figure}

In Fig.~\ref{fig:result}, we examine the recovery performance of the proposed E-MDNL technique as a function of the outlier ratio, which is defined as the ratio of the number of outliers to total number of observed distances. 
We observe that E-MDNL is more effective than the conventional MDNL in addressing the outlier problem. For example, we observe that E-MDNL achieves about 57\% reduction in the MSLE over MDNL at outlier ratio of 0.2.

%----------------------------------------------------------

%----------------------------------------------------------

\section{Conclusion}
In this paper, we proposed a novel DL and LRMC-aided localization
technique, called E-MDNL, to handle outliers in many practical IoT scenarios.
To address outliers effectively, we model then as a sparse matrix $\mathbf{L}$ and add a regularization term of $\mathbf{L}$ into the optimization problem.
By alternately updating $\mathbf{D}$ and $\mathbf{L}$ using closed-form solutions and updating $\mathbf{X}$ using DNN,
E-MDNL can accurately recover the outlier-contaminated Euclidean distance matrix. 
From the extensive simulations in various outlier scenarios, we confirm that
the proposed E-MDNL technique is effective in controlling the outliers. 
To acquire precise distance information without relying on RF signals, exploring IoT sensors that exploit visual sensing and computer vision~\cite{vomtc} is a promising future research direction.

 %The key idea of the proposed E-MDNL technique is to exploit the properties (low-rank, symmetry, and zero diagonal and positive non-diagonal entries) of a Euclidean distance matrix when recovering the desired Euclidean distance matrix $\mathbf{D}$. 
 %To this end, we first expressed $\mathbf{D}$ in a form of the sensor coordinate matrix $\mathbf{X}$ and then jointly recovered $\mathbf{D}$ and $\mathbf{X}$ by employing DNN.

% We hope our work will serve as a useful guide for researchers who want to use the VOMTC dataset in their CV-aided wireless application.
%To download VOMTC and the object detection train/test codes discussed in this paper, check out https://github.com/islab-github/VOMTC.

\vspace{-0.15cm}

% ======================================================================== %
\bibliographystyle{IEEEtran}

\end{document}